\documentclass[default,iicol]{sn-jnl}

\jyear{2023}%

\theoremstyle{thmstyleone}%
\theoremstyle{thmstyletwo}%
\theoremstyle{thmstylethree}%
\usepackage{overpic}
\usepackage[none]{hyphenat}

\usepackage{epsfig}
\usepackage{amsmath}
\usepackage{amssymb}
\usepackage{bm}
\usepackage{multirow}
\usepackage{subcaption}
\usepackage{hhline}
\usepackage{pifont}
\usepackage{wrapfig}

\newcommand{\Sa}[1]{§}

\def\ie{\emph{i.e.}}

\def\etal{{\em et al.}}

\newcommand\blfootnote[1]{%
  \begingroup
  \renewcommand\thefootnote{}\footnote{#1}%
  \addtocounter{footnote}{-1}%
  \endgroup
}

\newcommand{\cmark}{\ding{51}}
\newcommand{\xmark}{\ding{55}}

\newcommand{\best}[1]{\textcolor{red}{\textbf{#1}}}%
\newcommand{\sbest}[1]{\textcolor{blue}{\textbf{#1}}}%
\newcommand{\myPara}[1]{\vspace{.08in}\noindent\textbf{#1}\quad}

\begin{document}

\title[Crowd Counting]{Rethinking Global Context in Crowd Counting}


\author[1]{\fnm{Guolei} \sur{Sun}}

\author[2]{\fnm{Yun} \sur{Liu}$^\dagger$}
 
\author[3]{\fnm{Thomas} \sur{Probst}}

\author[1]{\fnm{Danda Pani} \sur{Paudel}}

\author[1]{\fnm{Nikola} \sur{Popovic}}

\author[1]{\fnm{Luc Van} \sur{Gool}}


\affil[1]{\orgdiv{Computer Vision Lab}, \orgname{ETH Zürich}, \orgaddress{\city{Zürich},  \country{Switzerland}}}

\affil[2]{\orgdiv{Institute for Infocomm Research}, \orgname{A*STAR},  \country{Singapore}}

\affil[3]{\orgdiv{Magic Leap}, \orgaddress{\city{Zurich},  \country{Switzerland}}}



\abstract{
This paper investigates the role of global context for crowd counting. Specifically, a pure transformer is used to extract features with global information from overlapping image patches. Inspired by classification, we add a context token to the input sequence, to facilitate information exchange with tokens corresponding to image patches throughout transformer layers. Due to the fact that transformers do not explicitly model the tried-and-true channel-wise interactions, we propose a token-attention module (TAM) to recalibrate encoded features through channel-wise attention informed by the context token. Beyond that, it is adopted to predict the total person count of the image through regression-token module (RTM).
Extensive experiments on various datasets, including ShanghaiTech, UCF-QNRF, JHU-CROWD++ and NWPU, demonstrate that the proposed context extraction techniques can significantly improve the performance over the baselines.
}

\keywords{crowd counting, vision transformer, global context, attention, density map}

\maketitle

\section{Introduction}
At first sight\blfootnote{$\dagger$ Corresponding authors.}, counting the size of a crowd present in an image is equivalent to the problem of  detecting and counting of person instances~\cite{ge2009marked,zhao2003bayesian}. Such direct approaches however have been shown not to perform well, because generic detectors suffer from the small instance size and severe occlusions present in crowded regions~\cite{liu2019context,hu2020count} -- typically a person covers only a small number of pixels,  and only few body parts are visible (often just the head)~\cite{idrees2018composition}. State-of-the-art crowd counting approaches therefore rely on the prediction of crowd density maps, a localized, pixel-wise measure of person presence~\cite{zhang2015cross,zhang2016single,8078491,sam2017switching,sam2018divide,li2018csrnet,ranjan2018iterative,idrees2018composition,shi2018crowd,cao2018scale,liu2019context,wang2019learning,shi2019revisiting,jiang2019crowd,zhang2019attentional,wan2019adaptive,shi2019counting,yan2019perspective,xiong2019open,liu2019crowd,ma2019bayesian,liu2019exploiting,jiang2020attention,liu2020weighing,wan2020modeling,wang2020DMCount}.

To this end, underlying network architectures need to integrate context across location and scales~\cite{liu2019context,li2018csrnet,ma2020learning}. This is crucial due to the vast variety of possible appearances of a given crowd density. In other words, the ability to integrate a large context makes it possible to adapt the density estimation to an expectation raised by the given scene, beyond the tunnel vision of local estimation.
\textit{Geometry} and \textit{semantics} are two of the main aspects of scene context \cite{lian2019density,kopaczewski2015method}, that can serve this goal for crowd counting \cite{lian2019density,He2020CPSPNetCC}.
Unfortunately, even if we manage to model and represent such knowledge, it is very cumbersome to obtain, and therefore not practical for many applications of image-based crowd counting. This also reflects the setup of the most popular crowd counting challenge datasets considered in this paper~\cite{zhang2016single,idrees2018composition, sindagi2020jhu, wang2020nwpu}.

On the bright side, even in the absence of such direct knowledge, we can benefit from the recent progress in geometric and semantic learning on a conceptual level -- by studying the inductive biases.
In fact, the development of computer vision in the last decade demonstrated the possibility to implicitly learn representations capturing rich geometric~\cite{watson2020learning} and semantic~\cite{SETR,sun2020mining} information from a single image.
Recently, the advantageous nature of global interaction over convolutional neural networks (CNNs) has been demonstrated for both geometric features for monocular depth prediction~\cite{yang2021transformers}, as well as for semantic features in segmentation~\cite{SETR,xie2021segformer}.
The aforementioned works attribute the success of the transformer \cite{vaswani2017attention,dosovitskiy2021image} to global receptive fields, which has been a bottleneck in previous CNN-based approaches. Moreover, CNNs by design apply the same operation on all locations, rendering it a sub-optimal choice for exploiting information about the geometric and semantic composition of the scene.

As geometric and semantic understanding are crucial aspects of scene context for the task of crowd counting, we hypothesize that superior capabilities of transformers on these aspects are also indicative of a more suitable inductive bias for crowd counting.
To investigate our hypothesis, we adapt the vision transformers~\cite{dosovitskiy2021image,yuan2021tokens,SETR} for the task of crowd counting.

Unlike image classification~\cite{dosovitskiy2021image}, crowd counting is a dense prediction task. Following our previous discussion, the learning of crowd counting is also predicated on the global context of the image. To capture both spatial information for dense prediction, as well as the necessary scene context, we maintain both local tokens (representing image patches) and a context token (representing image context). We then introduce a token attention module (TAM) to refine the encoded features informed by the context token. We further guide the learning of the context token by using a regression token module (RTM), that accommodates an auxiliary loss on the regression of the total count of the crowd. Following~\cite{SETR}, the refined transformer output is then mapped to the desired crowd density map using two deconvolution layers. 
Please refer to Fig.~\ref{fig:framework} for an illustration of the overall framework.

In particular, our proposed {TAM} is designed to address the observation that the multi-head self-attention (MHSA) in vision transformers only models spatial interactions, while the tried-and-true channel-wise interactions have also been proved to be of vital effectiveness \cite{hu2018squeeze,woo2018cbam}.
To this end, TAM imprints the context token on the local tokens by conditional recalibration of feature channels, therefore explicitly modelling channel-wise interdependencies.
Current widely-used methods to achieve this goal includes SENet~\cite{hu2018squeeze} and CBAM~\cite{woo2018cbam}. They use simple aggregation technique such as global average pooling or global maximum pooling on the input features to obtain channel-wise statistics (global abstraction), which are then used to capture channel-wise dependencies. For transformers, we propose a natural and elegant way to model channel relationships by extending the input sequence with a context token and introducing the TAM to recalibrate local tokens through channel-wise attention informed by the context token. The additional attention across feature channels further facilitates the learning of global context. 

We also adopt context token which interacts with other patch tokens throughout the transformers to regress the total crowd count of the whole image. This is achieved by the proposed RTM, containing a two-layer MLP. 
On the one hand, the syzygy of TAM and RTM forces the context token to collect and distribute image-level count estimates from and to all local tokens, leading to a better representation of context token.
On the other hand, it helps to learn better underlying features for the task and reduce overfitting within the network, similar to \textit{auxiliary-task learning}~\cite{gidaris2018unsupervised}.

In summary, we provide another perspective on density-supervised crowd counting, through the lens of learning features with global context. Specifically, we introduce a context token tasked with the refinement of local feature tokens through a novel framework of token-attention and regression-token modules. Our framework thereby addresses the shortcomings of CNNs with regards to capturing global context for the problem of crowd counting.
We conduct experiments on various popular datasets, including ShanghaiTech, UCF-QNRF, JHU-CROWD++ and NWPU.
The experimental results demonstrate that the proposed context extraction techniques can significantly improve the performance over the baselines and thus open a new path for crowd counting.

\section{Related works}
\subsection{Crowd Counting}
{Most crowd counting methods are based on convolutional neural networks (CNNs), which can be divided into three categories: counting by regression~\cite{chen2013cumulative,wang2015deep}, counting by detection~\cite{ge2009marked,li2008estimating,zhao2003bayesian}, and counting by estimating density maps~\cite{zhang2015cross,zhang2016single,8078491,sam2017switching,sam2018divide,li2018csrnet,ranjan2018iterative,idrees2018composition,shi2018crowd,cao2018scale,liu2019context,wang2019learning,shi2019revisiting,jiang2019crowd,zhang2019attentional,wan2019adaptive,shi2019counting,yan2019perspective,xiong2019open,liu2019crowd,ma2019bayesian,liu2019exploiting,jiang2020attention,liu2020weighing,wan2020modeling,wang2020DMCount,wang2021uniformity,song2021rethinking,liu2021exploiting}. The regression-based methods directly regress the total count of the crowd in the image, while the location of the people is not considered. Detection-based approaches first detect the people and then count the number of detections. However, those methods do not perform well in many interesting cases, where detection is difficult due to occlusions and high density of people. As a consequence, the mainstream direction of crowd counting is to estimate the density map of the image and then sum over the density map to obtain the total count. For this work, we also follow the direction of estimating density map. Different from existing methods, we target the crowd counting problem from the perspective of global information.}

{The methods~\cite{liu2019context,li2018csrnet,zhang2015cross} which exploit large receptive field for crowd counting have been proposed. The techniques include: using spatial average pooling~\cite{liu2019context} or dilation convolution~\cite{li2018csrnet}, and increasing network depth~\cite{zhang2015cross}. However, the receptive field is still limited, rather than global. Technically, only local information is used. In this work, we propose to use global information for crowd counting, by taking advantage of recent transformer technique. To the best of our knowledge, there are only limited works~\cite{liang2021transcrowd} adopting vision transformers to conduct crowd counting. However, the method of~\cite{liang2021transcrowd} is concerned with weakly supervised crowd counting in the sense of only regressing the total count, where dot annotations are not available. Its performance therefore cannot compete with the mainstream point-supervised crowd counting methods on most standard benchmarks~\cite{zhang2016single,idrees2018composition,wang2020nwpu}. Differently, we investigate point-supervised crowd counting using vision transformers and show the effectiveness of global context in crowd counting. }

\subsection{Vision Transformer}
The transformer, relying on self-attention mechanism~\cite{vaswani2017attention}, was first introduced in natural language processing~\cite{vaswani2017attention}, and has been dominating this area ever since. In general, a transformer contains a {MHSA} module and a multi-layer perceptron (MLP), to model the contextual information within input sequences through global interaction. Recently, pioneer works such as ViT~\cite{dosovitskiy2021image} and DETR~\cite{carion2020end} utilize transformers to solve vision problems. Transformers have shown to be effective in tasks of image classification~\cite{dosovitskiy2021image}, object detection~\cite{carion2020end}, and semantic/instance segmentation~\cite{SETR}.
{However, the exploration of transformers for crowd counting~\cite{liang2021transcrowd} has been limited. In this paper, we demonstrate the power of transformers in point-supervised crowd counting setup, where persons are represented with a dense map. }

\begin{figure*}[!t]
    \centering
    \includegraphics[width=0.999\linewidth]{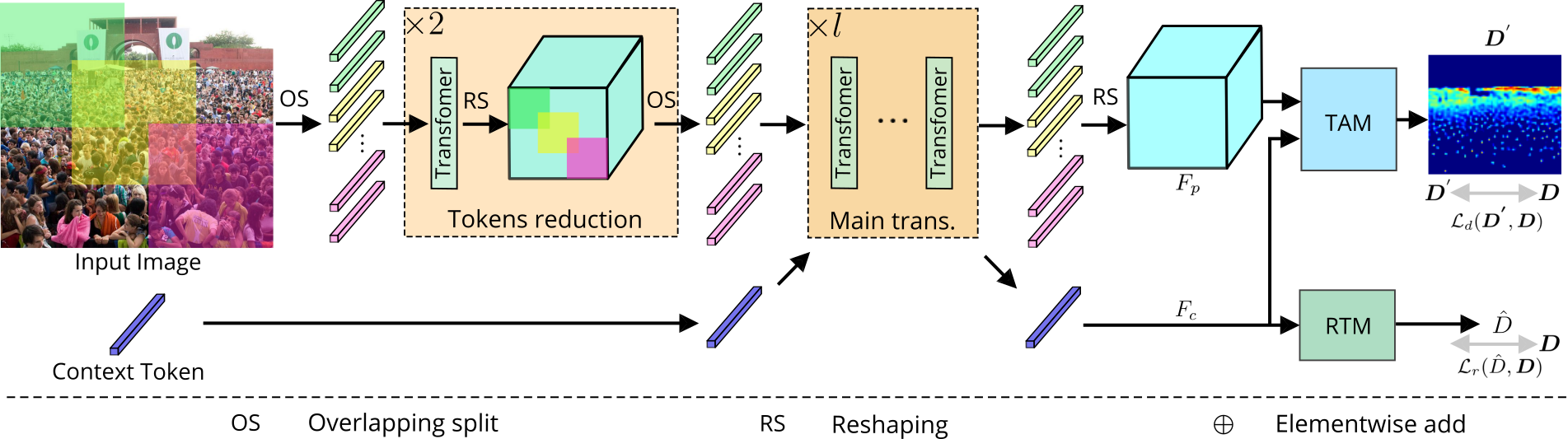}
\caption{\textbf{Network Overview.} The input image is first split into overlapping patches. Then, those patches go through tokens reduction block and main transformer to learn features with global information. To abstract global information, context token (blue vector) is added to the input sequence before the main transformer. The encoded features are processed by {TAM} and regression-token module (RTM). 
The small decoder after TAM is not shown for simplicity.
}
\label{fig:framework}
\end{figure*}

\section{Method} \label{sec:method}
\subsection{Problem Definition}
Given a training dataset of images $\{\bm{I}_i\}_{K} \subseteq \mathbb{R}^{ c \times h \times w}$ and crowd density label maps $\{\bm{D}_i\}_K \subseteq  \mathbb{R}^{h\times w}$, our goal is to learn a neural network model $\mathcal{M}: \mathbb{R}^{ c \times h \times w} \rightarrow \mathbb{R}^{h \times w}$, that estimates the crowd density map $\bm{D}^{'} \!=\! \mathcal{M}(\bm{I})$ and therefore counts the number of visible people $\lVert \bm{D}^{'}  \rVert_1$ from an unseen image $\bm{I}$.

\subsection{Transformer-based Crowd Counting}
Most crowd counting methods in the literature that consider crowd counting as a dense prediction task are based on CNNs~\cite{li2018csrnet,liu2019context,shi2018crowd}. 
Since CNN-based encoders can only exploit the local information within the fix-sized window, some approaches are proposed to increase the receptive fields, by dilated convolutions~\cite{li2018csrnet} or using deeper networks~\cite{wang2019learning}. 
In this section, we present our transformer-based approach for crowd counting, which is designed to overcome this limitation by explicitly modelling global context.
Our presentation follows the data flow of our framework as depicted in Fig.~\ref{fig:framework}. 

\myPara{Overlapping Split.} In the seminal ViT~\cite{dosovitskiy2021image}, the input image is split into non-overlapping patches, leading to the problem that the local structure around the patches is destroyed. Instead, we split the input into overlapping patches, following \cite{yuan2021tokens}. 
The process of overlapping split is similar to an convolution operation and the patch size of $k\times k$ is similar to the kernel size. Specifically, the input image $\bm{I}$ is first padded by $p$ pixels on each side. The overlapping patches are obtained by moving the patch window ($k\times k$) across the whole image with stride $s$ ($s<k$). Each patch has $k\times k\times c$ elements, which are flattened to $\mathbb{R}^{ck^2}$. The length of patches is given by
\begin{align}\small
\begin{split}
        N_0=h_0 \times w_0,
\end{split}
\end{align}
where $ h_0=\lfloor\frac{h+2p-k}{s}+1\rfloor$ and $w_0=\lfloor\frac{w+2p-k}{s}+1\rfloor$.
After concatenating all patches together, image tokens are obtained, denoted by $Z_0 \in \mathbb{R}^{N_0\times ck^2}$. Later, we process $Z_0$ by the tokens reduction block, followed by the main transformer.

\myPara{Tokens Reduction.} We first input $Z_0$ to a transformer layer and obtain $Z_1$, formulated as
\begin{align}\small
\begin{split}
        Z_1={\rm MLP}({\rm MHSA}(Z_0)),
\end{split}
\end{align}
where $Z_1 \in \mathbb{R}^{N_0\times d}$, and $d$ is the dimension of $query$, $key$, and $value$. Since the sequence length $N_0$ is relatively large due to the overlapping split, we reshape $Z_1$ back to $\mathbb{R}^{h_0 \times w_0 \times d}$ and perform overlapping split again to reduce the spatial size by stride $s$. Let $Z_1^{'}$ be the obtained tokens with size of $\mathbb{R}^{N_1\times dk^2}$, and $N_1=h_1 \times w_1$, 
where $ h_1=\lfloor\frac{h_0+2p-k}{s}+1\rfloor$ and $w_1=\lfloor\frac{w_0+2p-k}{s}+1\rfloor$. Following \cite{yuan2021tokens}, this process is repeated twice and we obtain $Z_2^{'} \in \mathbb{R}^{N_2\times dk^2}$, where $N_2=h_2\times w_2$. The length of sequence $N_2$ is thereby reduced to a manageable scale. Since the dependency among those pixels around the original non-overlapping split (as in ViT~\cite{dosovitskiy2021image}) is well-modelled, we fix the length of sequence as $N=N_2$ and do not reduce it further, in order to maintain both the representation capability and efficiency. After projecting $Z_2^{'}$ to $T\in \mathbb{R}^{N\times d}$, we process $T$ by deep-narrow ViT~\cite{dosovitskiy2021image}. 

\myPara{Context Token.}
Recall that we approach crowd counting as a dense prediction problem, and each patch token transforms local RGB input to a local density map prediction. Therefore, even though the patch tokens $T$ are in principle able to interact globally in ViT~\cite{dosovitskiy2021image}, our mode of dense supervision renders each token to be primarily concerned with its local region. In order to foster global information exchange without compromising capacity for local features, we delegate the collection of global context to a context token $t_{con}$. 
In contrast, previous transformer-based approaches to dense prediction~\cite{SETR} only employ local tokens without explicitly modelling the global context.
In our framework, the context token is the key input for the {TAM} as described in \Sa{}\ref{subsec:method_tam}, which disseminates the global context back to the local tokens. The local tokens therefore remain dedicated to their local predictions. 
In \Sa{}\ref{subsec:method_rtm} we explain how to guide the learning of context token through the RTM module. But first, we give a brief description of the main transformer of our framework.

\myPara{Main Transformer.}
The main transformer follow the same architecture as ViT~\cite{dosovitskiy2021image}, but have less channels in intermediate layers to reduce redundancy within original ViT model. As laid out above, we append the context token $t_{con}$ to the patch tokens $T$ to facilitate global-local interaction. Following \cite{dosovitskiy2021image}, position embedding $E$ is also added. The main transformer is denoted as follows:
\begin{align}\label{Eq:main_tranformers}\small
\begin{split}
        &T_0=[T; ~t_{con}]+E, ~~~E\in \mathbb{R}^{(N+1)\times d},\\
        &T_i^{'}={\rm MHSA}(T_{i-1})+T_{i-1},  ~~~i=1,...,l,\\
        &T_{i}={\rm MLP}(T_i^{'})+T_i^{'}, ~~~i=1,...,l.\\
\end{split}
\end{align}
Here, $l$ is the number of layers in the main transformer.
$T_l$ is the feature sequence from the last layer of transformers. It has global receptive fields which are effective for crowd counting task. Since a context token is added in the beginning, we split $T_l$ as follows
\begin{align}\label{Eq:split}\small
\begin{split}
        &F_{p}=T_{l}[:N], ~~F_c=T_{l}[N],\\
\end{split}
\end{align}
where $F_p \in \mathbb{R}^{N\times d}$ is the feature corresponding to image patches, and $F_c \in \mathbb{R}^d$ is the feature vector corresponding to context token $t_{con}$. To recover spatial structure, $F_p$ is reshaped to $\mathbb{R}^{d \times h_2\times w_2}$. $F_p$ is further refined by {TAM} to predict the density map and $F_c$ is used by the proposed regression-token module (RTM) to predict the overall count for the image .
\\

\subsection{Token-attention Module (TAM)} \label{subsec:method_tam}
The task of the TAM is to refine the local feature map $F_p$ used to predict the crowd density map, conditioned upon the context token feature $F_c$. This will infuse the global context information into the local density predictions. Before presenting the details of TAM, we give a brief analysis of the preceding transformer layers to motivate the proposed mechanism. 

\myPara{Spatial and Channel Attention.} Recall that the token $T_l$ is produced from $T_{l-1}$ by the last transformer layer, which performs the operations  
\begin{align}\label{Eq:diss_tam}\small
\begin{split}
        &T_l^{'}={\rm softmax}(\frac{T_{l-1}\bm{W}_Q(T_{l-1}\bm{W}_K)^{T}}{\sqrt{d}})T_{l-1}\bm{W}_V+T_{l-1},  \\
        &T_{l}={\rm MLP}(T_l^{'})+T_l^{'}, \\
\end{split}
\end{align}
where $T_{l-1}/T_{l}^{'}/T_{l}\in \mathbb{R}^{N\times d}$, and $\bm{W}_Q/\bm{W}_K/\bm{W}_V\in \mathbb{R}^{d\times d}$ are the learnable parameters for generating $(query,key,value)$. For simplicity of notation, {MHSA} is represented by a special case where a single self-attention (SA) operation is performed. We can see that a token $T_{l}^{'}[i]$ (corresponding to a specific image patch or the context token), is generated by a weighted summation of tokens $T_{l-1}$. 
Therefore, transformers are inherently equipped with spatial attention mechanism which pays more attention to the relevant spatial regions (tokens). However, the feature channel interdependencies are not explicitly modelled in the transformer operations~\eqref{Eq:diss_tam}. 
Explicitly modelling channel relationships, so that the network has the capability to focus on important feature channels, leads to enhanced features~\cite{hu2018squeeze,woo2018cbam}.
This is also confirmed by our experiments: while no improvements are obtained by adding spatial attention, introducing channel attention yields better predictions.
To this end, we introduce TAM as a mechanism to perform feature channel attention.  

\begin{figure*}[t]
  \centering
      \includegraphics[width=0.999\linewidth]{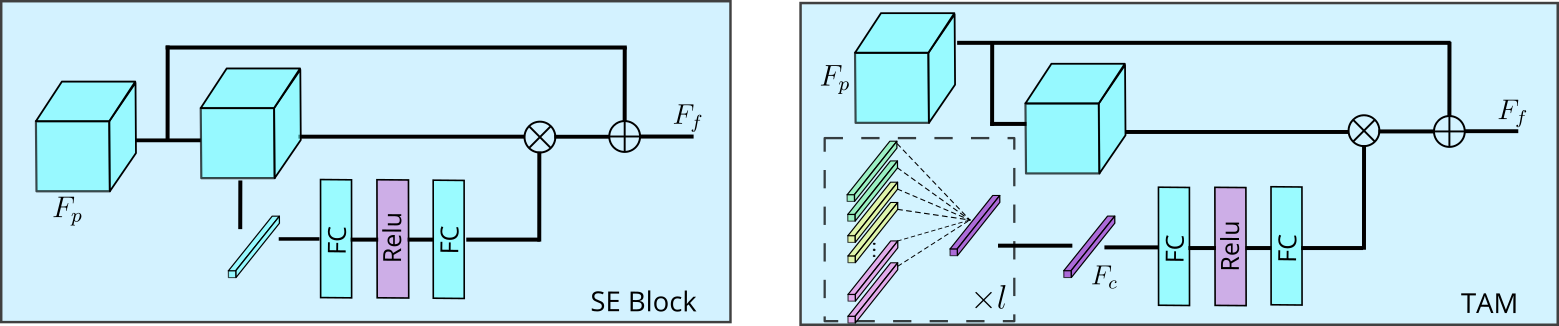}
\caption{\textbf{Comparison between SE block~\cite{hu2018squeeze} and TAM.} Different from SE block which obtains global information from input features, TAM adopts context token feature to provide channel relations.
}
\label{fig:se-tam}
\end{figure*}

\begin{figure}[t]
  \centering
      \includegraphics[width=0.7\linewidth]{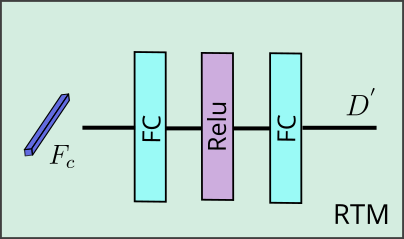}
\caption{The structure of RTM.}
\label{fig:rtm}
\end{figure}

\myPara{Global Abstraction.} Global abstraction is used to provide cues for channel interdependencies. In SE, global abstraction is obtained by conducting global average pooling across spatial dimensions on the input itself, while CBAM~\cite{woo2018cbam} merges both global average pooling and global maximum pooling. For transformers, we propose a natural and elegant approach to abstract global information, by extending the input sequence with a context token, as introduced above. Since the obtained feature $F_c$ from context token has a global overview throughout transformer layers, we adopt it to provide information on which channels are important for predicting density map. The comparison between SE and TAM is shown in Fig.~\ref{fig:se-tam}, where the $\rm sigmoid$ is omitted for simplicity. The superiority of the proposed TAM over SE~\cite{hu2018squeeze} and CBAM~\cite{woo2018cbam} is validated in \Sa{}\ref{exp:ablation_study}.

\myPara{Token-adaptive Recalibration.}
To capture channel-wise interactions, $F_c$ is projected by an MLP with $\rm ReLU$ activation, learning a weight vector $F_c^{'}$ which is used to re-weight the feature across the channels. $F_c^{'}$ is obtained by
\begin{align}\small
\begin{split}
        &F_c^{'}={\rm sigmoid}({\rm MLP}(F_c)).\\
\end{split}
\end{align}
Here, $F_c^{'}\in \mathbb{R}^{d}$ and ${\rm sigmoid}$ function is used to squeeze each element to a range of 0 and 1. We use a convolution layer to prepare $F_p$ for the re-weighting 
and obtain $F_p^{'}$. After the recalibration, we add a skip connection~\cite{he2016deep} with $F_p$ to derive the final feature map $F_f \in \mathbb{R}^{h_2\times w_2 \times d}$, like
\begin{align}\small
\begin{split}
        &F_f=F_p+F_p^{'}\otimes F_c^{'}.\\
\end{split}
\end{align}
Through the {TAM}, the network can increase sensitivity to informative features which are important for downstream processing. 

\subsection{Regression-token Module (RTM)}\label{subsec:method_rtm}
Recall that the context token is used to collect global context over the whole image.
It has a global overview of all image patches, through exchanging information with feature vector of each patch throughout all layers.
Therefore, we adopt $F_c$ to predict the overall count of people for the whole image. 
A two-layer MLP with $\rm ReLU$ activation is used to predict the total count $\hat{D}$, given by
\begin{align}\small
\begin{split}
        &\hat{D}={\rm MLP}(F_c).\\
\end{split}
\end{align}
The structure of RTM is shown in Fig.~\ref{fig:rtm}.
We use $L_1$ loss to reduce the difference between $\hat{D}$ and ground-truth count, as follows, 
\begin{align}\label{Eq:rtm-loss}\small
\begin{split}
        &\mathcal{L}_r(\hat{D}, \bm{D})=\vert\hat{D}-\lVert \bm{D} \rVert_1\vert.\\
\end{split}
\end{align}
Note that we only use this module during training, the predicted count for an image during test is obtained by summing over the predicted density map $\bm{D}^{'}$ (\Sa{}\ref{subsec:loss_func}), following other density-map based approaches~\cite{li2018csrnet,wang2020DMCount}. 
The benefits of RTM are two-fold. First, it forces to learn better context-token feature, which provides better information on the importance of each channel and enhance the final feature map $F_f$. Guiding the network to count the crowd through the context-token therefore encourages information exchange between context and patch tokens. In addition, it helps to learn better underlying feature representations and reduce over-fitting. This can be understood from a view of \textit{auxiliary-task learning}~\cite{gidaris2018unsupervised,odena2017conditional}, which has shown to be effective in segmentation~\cite{tian2020conditional,sun2020mining}.

\subsection{Density Map Prediction and Loss Functions}\label{subsec:loss_func}
To predict the density map $\bm{D}^{'}$, the feature map $F_f$ is processed by a decoder containing two convolutional layers.
We supervise the density map prediction using the distribution matching loss introduced by~\cite{wang2020DMCount}. To avoid dimension mismatch, the ground-truth $\bm{D}$ is resized to have the same size as $\bm{D}^{'}$, \ie, $\bm{D}\in \mathbb{R}^{h_2\times w_2}$. Specifically, the losses for learning the density map is a combination of counting loss, optimal transport loss~\cite{villani2008optimal} and variation loss, denoted as,
\begin{align}\label{Eq:loss_function}\small
\begin{split}
        \mathcal{L}_{d}(\bm{D}^{'},\bm{D})&=\vert\lVert \bm{D}^{'} \rVert_1-\lVert \bm{D} \rVert_1\vert+ \mathcal{L}_{OT}(\bm{D}^{'},\bm{D})\\
        &~~~~+ \mathcal{L}_{TV}(\bm{D}^{'},\bm{D}),\\
\end{split}
\end{align}
where $\mathcal{L}_{OT}$ is the optimal transport loss, and $\mathcal{L}_{TV}$ is the total variation loss. The first term measures the difference of the total count between the predicted density map and the ground-truth binary mask. 
The second (optimal transport loss) and third terms (variation loss) are used to minimize the distribution difference between $\bm{D}^{'}$ and $\bm{D}$ by regarding the density map as a probability distribution.
Please refer to \cite{wang2020DMCount} for more details. The total loss function for the proposed method is given by,
\begin{align}\label{Eq:final_loss}\small
\begin{split}
        &\mathcal{L}=\mathcal{L}_{d}(\bm{D}^{'},\bm{D})+\lambda_r \mathcal{L}_r(\hat{D}, \bm{D}),\\
\end{split}
\end{align}
where $\lambda_r$ is the weight for the regression loss $\mathcal{L}_r$ from \eqref{Eq:rtm-loss}. The final predicted count for inference is the summation over the predicted density map, given by $\lVert \bm{D}^{'} \rVert_1$.

\begin{table*}[!t]
\centering
\setlength{\tabcolsep}{6.0mm}
\resizebox{.999\textwidth}{!}{%
\begin{tabular}{l|c|c|c|c|c|c|c}
\hline
 & & \multicolumn{2}{c|}{ShanghaiTech A}                          & \multicolumn{2}{c|}{ShanghaiTech B}  & \multicolumn{2}{c}{~~~UCF-QNRF~~~}        \\ 
 \hhline{*{2}{|~}*{6}{|-}}
{\multirow{-2}{*}{Method}}  & {\multirow{-2}{*}{~Dot~}}  & \multicolumn{1}{c|}{~~MAE~~} & \multicolumn{1}{c|}{~~MSE~~} & \multicolumn{1}{c|}{~~MAE~~} & \multicolumn{1}{c|}{~~MSE~~} & \multicolumn{1}{c|}{~~MAE~~} & \multicolumn{1}{c}{~~MSE~~} \\ \hline \hline
Crowd CNN~\cite{zhang2015cross} & \cmark & 181.8 & 277.7 & 32.0  & 49.8  & -& - \\ \hline
MCNN~\cite{zhang2016single} & \cmark & 110.2 & 173.2 &  26.4 &  41.3 &277 & 426 \\ \hline
CMTL~\cite{8078491} & \cmark & 101.3 & 152.4 & 20.0 & 31.1 &252 & 514  \\ \hline
Switch CNN~\cite{sam2017switching} &\cmark & 90.4 & 135.0  & 21.6  & 33.4 & 228 & 445  \\ \hline
IG-CNN~\cite{sam2018divide}  &\cmark & 72.5 & 118.2  & 13.6  &21.1 & -& -  \\ \hline
CSRNet~\cite{li2018csrnet} & \cmark & 68.2 & 115.0 & 10.6 & 16.0    &-& -  \\ \hline
ic-CNN~\cite{ranjan2018iterative} &\cmark & 68.5 & 116.2  &  10.7 &16.0 &- & -  \\ \hline
CL-CNN~\cite{idrees2018composition} &\cmark & - & -  & -  &- &132 & 191  \\ \hline
SANet~\cite{cao2018scale} & \cmark & 67.0 &104.5 & 8.4&  13.6  & -& -   \\ \hline
CAN~\cite{liu2019context} &\cmark & 62.3 & 100.0&7.8  & 12.2 &107 &183  \\ \hline
SFCN~\cite{wang2019learning} &\cmark &64.8  &107.5 & 7.6  &  13.0  & 102& 171  \\ \hline
PACNN~\cite{shi2019revisiting} &\cmark & 62.4 & 102.0  & 7.6  &11.8 & -& -  \\ \hline
TEDnet~\cite{jiang2019crowd}  &\cmark & 64.2 &  109.1 &  8.2 & 12.8& 113.0& 188.0  \\ \hline

ANF~\cite{zhang2019attentional}  &\cmark & 63.9 & 99.4  & 8.3  &13.2 &110 & 174  \\ \hline
Wan \etal~\cite{wan2019adaptive} &\cmark & 64.7 & 97.1  & 8.1  &13.6 & 101&  176 \\ \hline
CFF~\cite{shi2019counting} &\cmark & 65.2 & 109.4  & {7.2}  &12.2 &- & -  \\ \hline
PGCNet~\cite{yan2019perspective} &\cmark & 57.0 & \sbest{86.0} & 8.8  &13.7 &- &  - \\ \hline
BL~\cite{ma2019bayesian} &\cmark & 62.8&101.8 & 7.7 &  12.7    & 88.7& 154.8   \\ \hline
L2R~\cite{liu2019exploiting} &\cmark & 73.6 & 112.0  & 13.7  & 21.4 & 124.0 & 196.0  \\ \hline
ASNet~\cite{jiang2020attention} &\cmark & 57.7 &  90.1  & - & - & 91.5  &159.7 \\ \hline
LibraNet~\cite{liu2020weighing}  &\cmark & \sbest{55.9} & 97.1  &  \best{7.3} & \best{11.3} & 88.1& \sbest{143.7}  \\ \hline
Yang \etal~\cite{yang2020weakly} & \xmark & 104.6 & 145.2  &  12.3 & 21.2 &- &  - \\ \hline

NoisyCC~\cite{wan2020modeling} &\cmark &61.9  &  99.6 & \sbest{7.4}  & \best{11.3} &85.8 & 150.6  \\ \hline
DM-Count~\cite{wang2020DMCount}   &\cmark & 59.7 & 95.7 & \sbest{7.4} & 11.8 & \sbest{85.6} & 148.3  \\ \hline 

MATT~\cite{lei2021towards} &\xmark & 80.1 & 129.4  & 11.7  &17.5 &- & -  \\ \hline\hline
Baseline & \cmark & 57.3 & 89.0 & \sbest{7.4}	 & 12.2  &85.7 & 150.8  \\ \hline
Ours & \cmark & \best{53.1} & \best{82.2} & \best{7.3} & \sbest{11.5}  & \best{83.4} & \best{143.4}  \\ \hline
\end{tabular}}
\caption{Comparison with state-of-the-art methods on ShanghaiTech A \cite{zhang2016single}, ShanghaiTech B \cite{zhang2016single}, and UCF-QNRF \cite{idrees2018composition} datasets. The best and second best results are shown in \best{red} and \sbest{blue}, respectively.}
\label{table:results_shanghai_qnrf}
\end{table*}

\section{Experiments}
We conduct extensive experiments on four benchmark crowd counting datasets~\cite{zhang2016single,idrees2018composition,sindagi2020jhu,wang2020nwpu} to validate the effectiveness of the proposed approach. We begin this section by 
introducing our experimental setting, followed by comparisons with previous methods. Finally, we perform ablation studies to examine the effectiveness of different components of our model.

\subsection{Experimental Setup}
\noindent\textbf{Implementation Details.} The number of layers $l$ in the main transformer is set to 14. We use the official T2T-ViT-14 model~\cite{yuan2021tokens} pretrained on ImageNet~\cite{russakovsky2015imagenet} for initialization. For data augmentation, we adopt random cropping and random horizontal flipping in all experiments. We use the Adam optimizer~\cite{kingma2014adam}, with learning rate and weight decay as 1e-5 and 1e-4, respectively. 
Following \cite{SETR}, we compute auxiliary losses at transformer layers $T_5$, $T_8$, and $T_{11}$, to provide intermediate supervision during training while only output from last layer is used for prediction. Our method is implemented in the PyTorch framework~\cite{paszke2019pytorch}, and experiments are conducted on a single NVIDIA Tesla GPU. We will release our implementation for reproducibility.

\myPara{Datasets.}
Experiments are conducted on four challenging datasets: ShanghaiTech~\cite{zhang2016single}, UCF-QNRF~\cite{idrees2018composition}, JHU-CROWD++~\cite{sindagi2020jhu} and NWPU~\cite{wang2020nwpu}. ShanghaiTech contains 1,198 images with  330,165 annotations, and UCF-QNRF has 1,535 images with more than one million counts. JHU-CROWD++ and NWPU are two largest-scale and most challenging crowd counting benchmarks.
JHU-CROWD++ consists of 4,822 images from diverse scenes with more than 1.5 million dot annotations, and NWPU contains 5,109 images with more than two million annotations. The results for the test set are obtained from the evaluation server.

\myPara{Evaluation Metrics.}
Following previous works~\cite{li2018csrnet,wang2020DMCount,liu2019context}, we use mean average error (MAE) and mean square error (MSE) to evaluate the counting performance. For NWPU dataset, we also use mean normalized absolute error (NAE) as evaluation metric, following \cite{wang2020nwpu,wang2020DMCount}.

\begin{table*}[!t]
\centering
\setlength{\tabcolsep}{4.0mm}
\resizebox{0.999\textwidth}{!}{%
\begin{tabular}{l|c|c|c|c|c|c} \hline
 & & & \multicolumn{2}{c|}{~~~Val~~~}                          & \multicolumn{2}{c}{~~~Test~~~}                      \\ 
 \hhline{*{3}{|~}*{4}{|-}}
{\multirow{-2}{*}{Method}} & {\multirow{-2}{*}{~~~Publication~~~}}  & {\multirow{-2}{*}{~~~Dot~~~}}  & \multicolumn{1}{c|}{~~~~MAE~~~~} & \multicolumn{1}{c|}{~~~~MSE~~~~} & \multicolumn{1}{c|}{~~~~MAE~~~~} & \multicolumn{1}{c}{~~~~MSE~~~~}  \\ \hline \hline
MCNN~\cite{zhang2016single}  & CVPR16 & \cmark & 160.6 & 377.7 & 188.9  & 483.4    \\ \hline
CMTL~\cite{8078491}  & AVSS17 & \cmark & 138.1 & 379.5 & 157.8  & 490.4    \\ \hline
SANet~\cite{cao2018scale}  & ECCV18 & \cmark & 82.1 &272.6 & 91.1 & 320.4     \\ \hline
CSRNet~\cite{li2018csrnet}  & CVPR18 & \cmark & 72.2 & 249.9 & 85.9 &  309.2    \\ \hline
CAN~\cite{liu2019context} & CVPR19 &\cmark & 89.5 & 239.3 & 100.1 & 314.0 \\ \hline
SFCN~\cite{wang2019learning} & CVPR19 &\cmark & 62.9 &247.5 &  77.5 & 297.6    \\ \hline
BL~\cite{ma2019bayesian} & ICCV19  &\cmark & 59.3 & 229.2 & 75.0 &  299.9      \\ \hline
MBTTBF~\cite{sindagi2019multi} & ICCV19  &\cmark & 73.8 & 256.8  & 81.8  & 299.1     \\ \hline
CG-DRCN~\cite{sindagi2020jhu} & PAMI20  &\cmark &57.6 &244.4 & 71.0 &  278.6      \\ \hline  \hline
Baseline & -&\cmark & \sbest{47.6} & \sbest{208.5} & \sbest{58.4} & \sbest{232.7}   \\ \hline
Ours & -&\cmark & \best{46.5} & \best{198.6} & \best{54.8} & \best{208.5}   \\ \hline
\end{tabular}}
\caption{Comparison with state-of-the-art methods on the JHU-CROWD++ dataset \cite{sindagi2020jhu}.}
\label{table:results_jhu}
\end{table*}

\begin{table*}[!t]
\centering
\setlength{\tabcolsep}{4.0mm}
\resizebox{0.999\textwidth}{!}{%
\begin{tabular}{l|c|c|c|c|c|c|c} \hline
 & & & \multicolumn{2}{c|}{~~~Val~~~}                          & \multicolumn{3}{c}{~~~Test~~~}                      \\ 
 \hhline{*{3}{|~}*{5}{|-}}
{\multirow{-2}{*}{Method}} & {\multirow{-2}{*}{~~~Publication~~~}}  & {\multirow{-2}{*}{~~~Dot~~~}}  & \multicolumn{1}{c|}{~~~MAE~~~} & \multicolumn{1}{c|}{~~~MSE~~~} & \multicolumn{1}{c|}{~~~MAE~~~} & \multicolumn{1}{c|}{~~~MSE~~~} & \multicolumn{1}{c}{~~~NAE~~~} \\ \hline \hline
MCNN~\cite{zhang2016single}  & CVPR16 & \cmark & 218.5 &700.6 & 232.5  &714.6  &  1.063 \\ \hline
CSRNet~\cite{li2018csrnet}  & CVPR18 & \cmark &104.8 &433.4 & 121.3 &  387.8  & 0.604 \\ \hline
CAN~\cite{liu2019context} & CVPR19 &\cmark &93.5 &489.9 & 106.3 &386.5 &0.295\\ \hline
SFCN~\cite{wang2019learning} & CVPR19 &\cmark &95.46 & 608.32&  105.7 & 424.1  & 0.254  \\ \hline
BL~\cite{ma2019bayesian} & ICCV19  &\cmark &93.64 & 470.38& 105.4 &  454.2    & 0.203  \\ \hline
KDMG~\cite{wan2020kernel} & PAMI20 &\cmark &- &- & 100.5& 415.5 & - \\ \hline
NoisyCC~\cite{wan2020modeling} & NeurIPS20 &\cmark &- &- &  96.9  &  534.2 & -  \\ \hline
DM-Count\cite{wang2020DMCount}   &  NeurIPS20 &\cmark & 70.5 &357.6   & 88.4  & 388.6 & \sbest{0.169} \\ \hline \hline
Baseline & -&\cmark & \sbest{69.0} & \sbest{314.0} & \sbest{86.6} & \best{359.1}  & 0.172\\ \hline
Ours & -&\cmark & \best{53.0} & \best{170.3} & \best{82.0} &  \sbest{366.9} & \best{0.164}\\ \hline
\end{tabular}}
\caption{Comparison with state-of-the-art crowd counting methods on the NWPU dataset \cite{wang2020nwpu}.}
\label{table:results_nwpu}
\end{table*}

\subsection{Crowd Counting Results}\label{sec:exp_results}
\noindent\textbf{Baseline.}
The baseline model is based on the same transformers, also adopting loss functions~\eqref{Eq:final_loss}, without using {TAM} and regression-token module (RTM). 

\myPara{Quantitative Comparisons.} 
{For comparisons, we choose mainstream and popular methods. They can be divided into three groups. The VGG-based approaches include CSRNet~\cite{li2018csrnet}, ic-CNN~\cite{ranjan2018iterative}, CAN~\cite{liu2019context}, PACNN~\cite{shi2019revisiting}, Wan \etal~\cite{wan2019adaptive}, PGCNet~\cite{yan2019perspective}, BL~\cite{ma2019bayesian}, L2R~\cite{liu2019exploiting}, ASNet~\cite{jiang2020attention}, LibraNet~\cite{liu2020weighing}, Yang \etal~\cite{yang2020weakly}, NoisyCC~\cite{wan2020modeling}, DM-Count~\cite{wang2020DMCount}, MATT~\cite{lei2021towards}, and MBTTBF~\cite{sindagi2019multi}. The ResNet-based methods include SFCN~\cite{wang2019learning} and CG-DRCN~\cite{sindagi2020jhu}. Other CNN-based algorithms include Crowd CNN~\cite{zhang2015cross}, MCNN~\cite{zhang2016single}, CMTL~\cite{8078491}, Switch CNN~\cite{sam2017switching}, IG-CNN~\cite{sam2018divide}, CL-CNN~\cite{idrees2018composition}, SANet~\cite{cao2018scale}, TEDnet~\cite{jiang2019crowd}, ANF~\cite{zhang2019attentional}, CFF~\cite{shi2019counting}.}

The comparisons with other methods on various datasets are shown in Table~\ref{table:results_shanghai_qnrf}, Table~\ref{table:results_jhu} and Table~\ref{table:results_nwpu}. For all datasets, the proposed method performs favorably. It shows that our approach is stable across different datasets.
The reported tables show that our baseline is already comparable to the state-of-the-art CNN-based methods. Notably, our full model outperforms the baseline model in almost all experiments, which validates the effectiveness of the proposed modules.
In all cases, our method significantly outperforms DM-count~\cite{wang2020DMCount}, although both methods use the same decoder and loss functions to learn the density map. For example, when compared with DM-count on ShanghaiTech~A \cite{zhang2016single}, our model reduces MAE from 59.7 to 53.1, and MSE from 95.7 to 82.2. This demonstrates the importance of global context features for the task of crowd counting. 

On two largest-scale and most challenging benchmarks such as JHU-CROWD++~\cite{sindagi2020jhu} and NWPU~\cite{wang2020nwpu}, our approach significantly outperforms the previous best results. More specifically, our method improves BL~\cite{ma2019bayesian}, the best method on JHU-CROWD++ test set, by reducing MAE from 75.0 to 54.8 and MSE from 299.9 to 208.5. Similarily on NWPU dataset, our method outperforms DM-count~\cite{wang2020DMCount}, the best method on the NWPU test set, by a margin of 6.4 and 21.7 on MAE and MSE, respectively. Note that the annotations for the NWPU test set are not publicly available and the corresponding results are obtained from the evaluation server.

\myPara{Computing Time.} {Using an input image with size $256 \times 256$ and a Nvidia RTX 6000 GPU, the computing time of our method is 21.47 milliseconds while the time for DM-count~\cite{wang2020DMCount} is 16.98 milliseconds. Note that DM-count is currently state-of-the-art CNN-based method and our method is based on transformer. Because of the use of transformer to establish global relation between features, our method consumes more time compared to CNN-based method. Developping more efficient crowd counting approach while using global information will be our future work.}

\begin{figure*}[t]
  \centering
  \includegraphics[width=.999\linewidth]{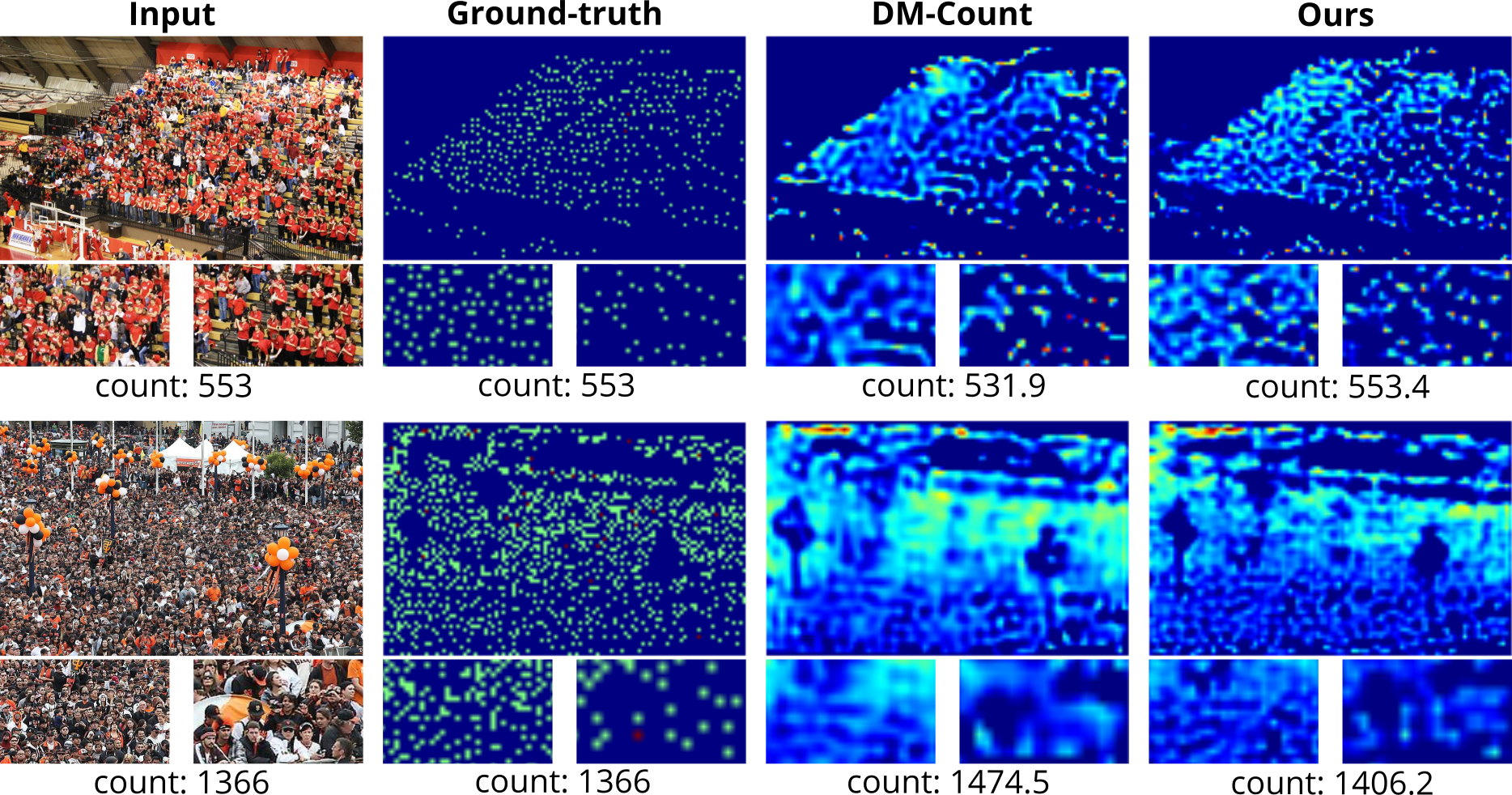}
\caption{\textbf{Density Map Visualization}. We compare the ground-truth density map, predicted density map from DM-count~\cite{wang2020DMCount} and the proposed method. Our approach produces better density map for both dense and sparse regions, leading to more accurate count predictions.}
\label{fig:qual}
\end{figure*}

\myPara{Visualizations.} Qualitative results of the predicted density maps are shown in Fig.~\ref{fig:qual}. Our method generates sharper density maps and exhibits better localization ability, compared to DM-count~\cite{wang2020DMCount}.

\begin{figure}[t]
  \centering
  \includegraphics[width=.99\linewidth]{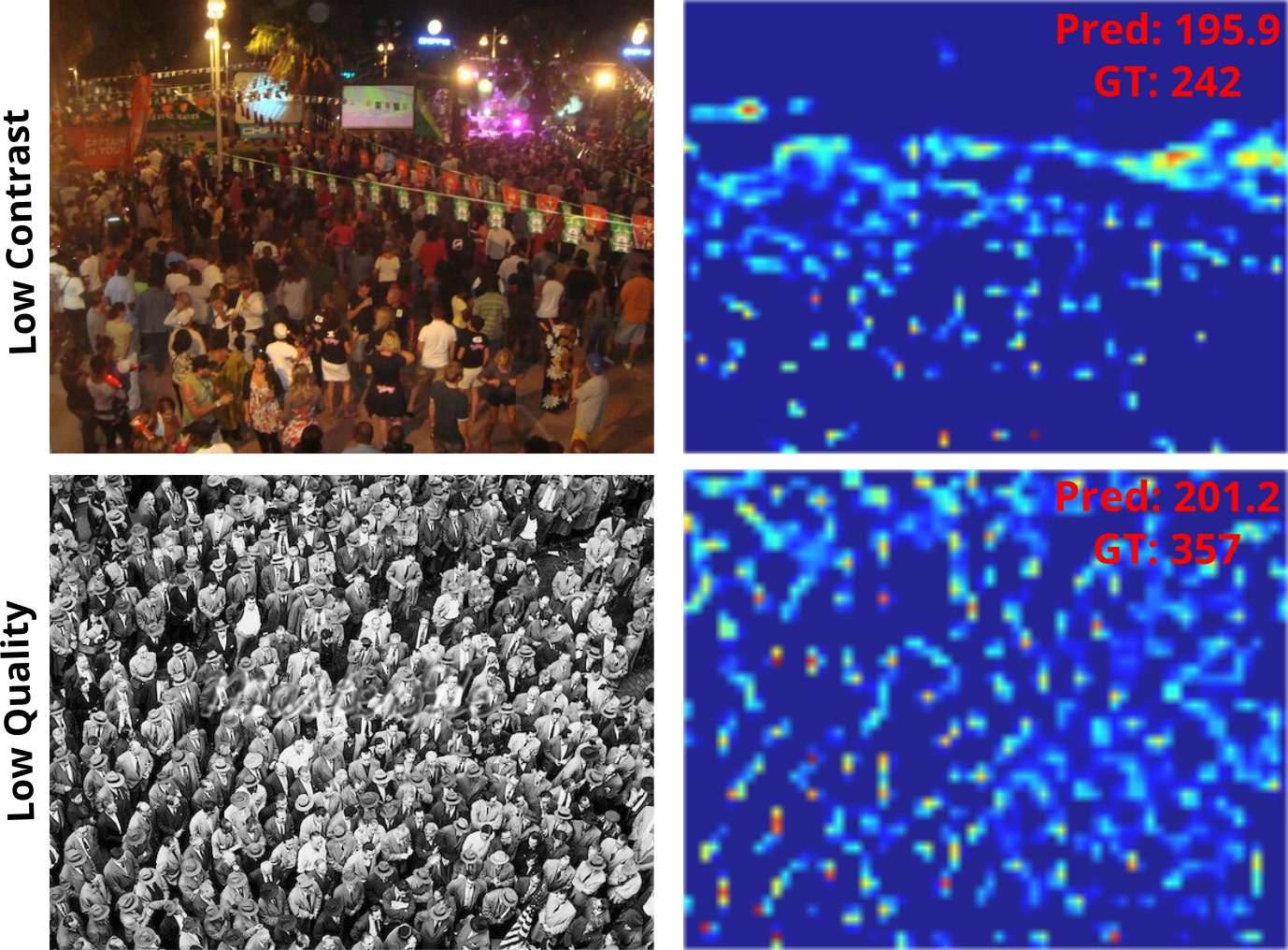}
\caption{Failure Cases. The failure cases are caused by the \textit{low contrast} or \textit{low quality} of the input images.}
\label{fig:failure}
\end{figure}

\begin{table*}[!t]
\centering
\begin{subtable}{0.440\linewidth}
\centering
\setlength{\tabcolsep}{4.0mm}
\captionsetup{font={normalsize}}
\resizebox{\textwidth}{!}{
	\begin{tabular}{l|c|c}
		\hline
{Method}  & ~~~~MAE~~~~ & ~~~~MSE~~~~ \\  \hline \hline
Baseline &  57.3 & 89.0 \\
Baseline+SE~\cite{hu2018squeeze}  & 55.9  & 87.7 \\ 
Baseline+CBAM~\cite{woo2018cbam}  & 56.2 & 90.0 \\ 
Baseline+TAM  & 55.1 & 86.2 \\
{Baseline+SE+RTM}~\cite{hu2018squeeze}  & 54.6  & 84.2 \\ 
Baseline+TAM+RTM & \textbf{53.1} & \textbf{82.2}\\ \hline
\end{tabular}}
\caption{}
\label{results:comp}
\end{subtable}%
\hspace{0.3in}
\begin{subtable}{0.455\linewidth}
\centering
\setlength{\tabcolsep}{3.5mm}
\captionsetup{font={normalsize}}
\resizebox{\textwidth}{!}{
	\begin{tabular}{l|c|c|c}
		\hline
{Method~~} & $\lambda_r$ & ~~~~MAE~~~~ & ~~~~MSE~~~~ \\  \hline \hline
\multirow{5}{*}{Ours} & 0.01& 53.2 & 83.6 \\ 
 &0.1 & \textbf{53.1} & \textbf{82.2}\\
 &0.2 & 53.2  & 82.4\\ 
 &0.5 & 53.3 & 82.4\\ 
  &1.0 & 54.0 & 82.6 \\ \hline
\end{tabular}}
\caption{}
\label{results:lambda}
\end{subtable}
\caption{Ablation study on \textbf{(a)} key components of our method and \textbf{(b)} $\lambda_r$ on ShanghaiTech A.}
\label{table:ablation_tam_rtm}
\end{table*}

\subsection{Ablation Study}\label{exp:ablation_study}
Following previous works~\cite{li2018csrnet,liu2019context,jiang2020attention,liu2020weighing}, 
we conduct ablation experiments on ShanghaiTech~A~\cite{zhang2016single}, to show the contributions of the key components of our method.

\myPara{TAM and RTM.} Table~\ref{results:comp} shows that the  token-attention module and regression-token module provide complementary improvements over baseline. Specifically, by adding TAM to the baseline, we observe an improvement of 2.2 in MAE and of 2.8 in MSE. 
The best results are achieved by combining TAM with RTM, resulting in an improvement of 4.2 MAE and 6.8 in MSE over the baseline.

\myPara{TAM vs. SE/CBAM.} We also compare the proposed TAM block with a SE block~\cite{hu2018squeeze} and a CBAM block~\cite{woo2018cbam}. The main difference between TAM and SE/CBAM is that the attention weight for TAM is obtained from context token, while SE/CBAM use feature itself to generate attention weight. As shown in Table~\ref{results:comp}, TAM outperforms SE/CBAM, demonstrating that context token contains better information to recalibrate features along channels. The result for CBAM which uses both channel and spatial attention shows that additionally adding spatial attention does not help feature learning, since transformers are naturally equipped with spatial attention, as hypothesized in \Sa{}\ref{subsec:method_tam}.

\myPara{Sensitivity Analysis.} Table~\ref{results:lambda} shows the results when varying $\lambda_r$ controlling the contribution of $\mathcal{L}_{r}$ in Equation~\eqref{Eq:final_loss}. 
We observe that our network is very robust to the choice of the $\lambda_r$ parameter.

\subsection{Failure Cases and Limitation}
\myPara{Failure Cases.}
{While our method achieves promising results on several datasets, there are cases where it does not perform well. We showed failure cases in Fig.~\ref{fig:failure}. When the input images have low contrast or low quality, which do not frequently appear in the training set, our method does not predict the similar people count as the ground truth.}

\myPara{Limitation.}
This paper aims at exploiting the effect of global context in crowd counting. Although we have achieved this goal by demonstrating the effectiveness of the proposed context extraction techniques, we do not explore how to incorporate our techniques into existing state-of-the-art counting methods \cite{liu2020weighing,wang2020DMCount,wan2020modeling,lei2021towards} for performance boosting. We leave this as the future work as this is totally about engineering. Moreover, we find that the improvement of our context techniques on large datasets \cite{sindagi2020jhu,wang2020nwpu} is much more significant than that on small datasets \cite{zhang2016single,idrees2018composition}. This may be a platitude that deep learning needs large-scale data to evaluate its real performance. Hence, we suggest researchers paying more attention to recent large-scale datasets \cite{sindagi2020jhu,wang2020nwpu} in the future.

\section{Conclusion}
In this paper, we study the value of global context information in crowd counting with point supervision. We build a strong baseline using transformers to encode features with global receptive fields. Based on that, we proposed two novel modules: token-attention module and regression-token module. Extensive experiments are conducted to validate the effectiveness of the proposed techniques. Our context techniques achieve significant improvement on ShanghaiTech \cite{zhang2016single}, UCF-QNRF \cite{idrees2018composition}, JHU-CROWD++ \cite{sindagi2020jhu}, and NWPU \cite{wang2020nwpu} datasets. Therefore, we conclude that facilitating the representation of global context significantly benefits crowd counting.

\bibliographystyle{IEEEtran}
\bibliography{reference}

\vspace{20pt}
\noindent{\bf Guolei Sun} 
is a PhD candidate at ETH Zurich, under supervision of Prof. Luc Van Gool. He received master degree in computer science from King Abdullah University of Science and Technology in 2018. From 2018 to 2019, he worked as a research engineer at the Inception Institute of Artificial Intelligence, UAE. His research interests lie in computer vision and deep learning for tasks such as semantic segmentation, video understanding, and object counting. He has published more than 20 papers in top journals and conferences such as TPAMI, CVPR, ICCV, and ECCV.

E-mail: sunguolei.kaust@gmail.com

ORCID iD: 0000-0001-8667-9656

\vspace{20pt}
\noindent{\bf Yun Liu} received his BEng and Ph.D. degrees from
Nankai University in 2016 and 2020, respectively. Then, he worked with Prof. Luc Van Gool for one and a half years as a postdoctoral scholar at
Computer Vision Lab, ETH Zurich. Currently, he is a senior scientist at Institute for Infocomm Research (I2R), A*STAR. His research interests include computer vision and machine learning..

E-mail: VAGRANTLYUN@gmail.com

ORCID iD: 0000-0001-6143-0264

\vspace{20pt}
\noindent{\bf Thomas Probst} received his master and PhD degrees from Ulm University and ETH Zurich, respectively. After that, he was a Postdoc researcher at the Computer Vision Lab under Prof. Luc Van Gool at ETH Zurich. The main focus of his research is on deep learning for geometry problems, and the perception of humans for robotics.

E-mail: thomas.tpr.probst@gmail.com

\vspace{20pt}
\noindent{\bf Danda Pani Paudel} is a researcher at the Computer Vision Lab, ETH Zurich, working with Prof. Luc Van Gool. His research interests include Computer Vision, Visual-SLAM, Unsupervised Learning, and Optimization Methods. He received his Doctoral degree (PhD) in 2015, and a Master's degree in computer vision in 2012, from University of Bourgogne, France. He also worked as a research scholar at University of Strasbourg, France, from 2013 to 2015, while devising global and local methods for 2D-3D registration problems.

E-mail: paudel@vision.ee.ethz.ch

\vspace{20pt}
\noindent{\bf Nikola Popovic} is a PhD candidate at the Computer Vision Lab at ETH Zurich (Switzerland). His research interests include computer vision, machine learning, and artificial intelligence. He has published a number of papers in major computer vision conferences.

E-mail: nikola.popovic@vision.ee.ethz.ch

\vspace{20pt}
\noindent{\bf Luc Van Gool} 
received the degree in electromechanical engineering at 
the Katholieke Universiteit Leuven in 1981. 
Currently, he is a professor at the Katholieke 
Universiteit Leuven in Belgium and the ETH in Zurich, Switzerland. 
He leads computer vision research at both places, 
and also teaches at both. He has been a program
committee member of several major computer vision conferences. His main interests include 3D reconstruction and modelling, object recognition, tracking, and gesture analysis, and the
combination of those. 
He received several Best Paper awards, won 
a David Marr Prize and a Koenderink Award, 
and was nominated Distinguished Researcher by 
the IEEE Computer Science committee. 
He is a co-founder of 10 spin-off companies.

E-mail: vangool@vision.ee.ethz.ch

\end{document}